\documentclass{article}

\usepackage[export]{adjustbox}
\usepackage{neurips_2022}
\usepackage{graphicx}
\def\BibTeX{{\rm B\kern-.05em{\sc i\kern-.025em b}\kern-.08em
    T\kern-.1667em\lower.7ex\hbox{E}\kern-.125emX}}

\usepackage[utf8]{inputenc} 
\usepackage[T1]{fontenc}    
\usepackage{hyperref}       
\usepackage{url}            
\usepackage{booktabs}       
\usepackage{amsfonts}       
\usepackage{nicefrac}       
\usepackage{microtype}      
\usepackage{xcolor}         
\usepackage{amsmath,amssymb,amsfonts}
\usepackage{algorithmic}
\usepackage{newunicodechar}

\title{Determination of toxic comments and unintended model bias minimization
using Deep learning approach}

\begin{document}

\maketitle

\begin{abstract}
  Online conversations can be toxic and subjected to threats, abuse, or harassment. To identify toxic text comments, several deep learning and machine learning models have been proposed throughout the years. However, recent studies demonstrate that because of the imbalances in the training data, some models are more likely to show unintended biases including gender bias and identity bias. In this research, our aim is to detect toxic comment and reduce the unintended bias concerning identity features such as race, gender, sex, religion by fine-tuning an attention based model called BERT(Bidirectional Encoder Representation from Transformers). We apply weighted loss to address the issue of unbalanced data and compare the performance of a fine-tuned BERT model with a traditional Logistic Regression model in terms of classification and bias minimization. The Logistic Regression model with the TFIDF vectorizer achieve 57.1\% accuracy, and fine-tuned BERT model's accuracy is 89\%. Code is available at \url{https://github.com/zim10/Determine_Toxic_comment_and_identity_bias.git}.
  
\end{abstract}

\section{Introduction}

Computers and cellphones are becoming necessary tools for doing daily tasks. People are increasingly using social media to communicate, using sites like Twitter and Facebook to share their opinions. Online conversations using those social media can be toxic and subjected to threats, abuse, or harassment. Toxic comments are the comments that are rude and disrespectful and are more likely to make someone leave a discussion \cite{risch2020toxic}. According to a 2020 Pew Research survey \cite{lewis2020online}, 41\% of internet users had experienced online harassment, with 18\% of those incidents being serious harassment.

It is hard to manually monitor and regulate toxic conduct in social media platforms and chat systems due to the volume of text discussions they include. Here NLP (Natural Language Processing) is a discipline that allows us to create models that can identify and categorize toxic comment. But Recent studies show that toxic comments classifiers often suffer from unintended bias, especially the bias towards “identity terms”, which are words or terms referring to people with specific demographic characteristics, such as ethnic origin, religion, gender, or sexual orientation \cite{Ji}. For example, \textbf{“You are a good woman”} or \textbf{“I am a gay woman”} or \textbf{“I am a black man”} is predicted as \textbf{“toxic”} comment. It has happened because the models have been extensively trained with certain keywords (such as \textbf{"black," "Muslim," "feminist," "woman," "gay,"} etc.) that are frequently used in harmful comments. As a result, if any of these keywords are used in the context of a comment that is not harmful, the model's bias towards those keywords will cause it to predict that the comment is toxic.

Therefore, the objective is to detect toxic comments and reduce the unintended bias concerning identity features such as race, gender, sex, religion, etc. using deep learning approach.
In summary, the main contributions of this research are as follows.

\begin{itemize}
    \item  Exploratory data analysis{(EDA)} to know the trends, patterns, and relationships among variable in the Dataset.
    \item Tuned the BERT and logistic regression models to classify toxic comments while reducing identity bias.
    \item Analyzes and compare the bias minimization performance and classification performance of the trained model.
\end{itemize}

\subsection{Research Questions}
In this research we will answer the following research questions.

\textbf{RQ1.} What metrics are best suitable to measure the bias minimization performance
of the logistics regression and Fine-tuned Bert models and why?

\textbf{Motivation:} In addition to classifying toxicity, the suggested approach(Fine tuned BERT) should reduce the identification bias towards commonly attacked non-toxic identities. A model is considered to have identity bias if its results vary when comments are associated to several identity groups \cite{risch2020toxic}. For instance,
consider the comments \textbf{“I’m proud Woman”} and \textbf{“I’m a proud gay woman”}, an ideal model should give a same toxic score to both the comments, whereas a biased model will generate different scores. 
The model's consistency across many identities will indicate whether the model is biased or not. So, a proper evaluation metric is need to choose that can be used to assess how consistent the model is across various identity groups. In order to select the metric that best matches this study challenges, we would like to review the numerous metrics that have been employed in the past.

\textbf{RQ2.} How well the logistic regression and Fine-tuned Bert models can measure the toxicity by reducing the unintended bias over certain identity terms?

\textbf{Motivation:} We should evaluate the models' performance after training them to find out how effectively they can detect toxicity in text conversation and how much unintentional bias they introduce toward often attacked non-toxic identity phrases.

\section{Related works}
\label{gen_inst}

Online user-generated content such as hate speech, harsh language, and cyberbullies are all examples of what is often referred to as toxic comments \cite{Irene}. Most studies treat Toxic comment classification as an issue of binary categorization. A few studies define toxic comment as multi-class classification tasks where a remark is allocated to one of the many different categories of toxic comments, or as multi-label tasks where a comment may be assigned to none, one, or many different types of toxic comments \cite{Betty}. Some studies have explored the nuances of toxic comments through sentiment analysis, delving into the emotional context and tone of the text to understand the severity of toxicity \cite{risch2020toxic}. Additionally, efforts have been made to incorporate contextual information and user behavior patterns into toxic comment detection, aiming to enhance the accuracy of identifying toxic content \cite{brassard2019subversive}. Furthermore, there is a growing interest in the development of cross-lingual toxic comment classification models to address the challenges posed by multilingual online platforms \cite{song2021study}.

Early research on Toxic comment classification  primarily makes use of traditional machine learning algorithms, such as logistic regression, Support Vector Machines and Naïve Baye \cite{Anna}. Many researchers are using machine learning and deep learning to provide automated answers for issues that arise in the real world. Using deep learning models to find toxic comments is one of these use cases. But recent research shows that some models tend to exhibit unintended biases (like gender, race, religion) due to the imbalances in the training data. The paper \cite{Lucas} looks at the issue of data imbalance problem that leads model to identity bias terms. They found that identity phrases are found significantly more frequently in toxic than in non-toxic comments, and they suggest that this imbalance causes the classifier to generalize identity terms as indicative features for toxicity and over-predict false positives. They manually add non-toxic comments containing identification keywords to equalize the proportion of these terms across toxic and non-toxic comments to remedy this issue.Irsoy et al. \cite{mehrabi2021survey}discusses various biases that were found
in a wide range of machine learning and deep learning models which were used in
any applications ranging from image recognition, text classification and so on.

Despite a substantial amount of research on toxic comment classification, it has been demonstrated that existing research focus on biased predictions based on identity terms. In this study, we fine tuned BERT model using transfer learning concept and  answer the above research question based on a data set and also train one traditional machine learning model (Logistic regression) to compare models performance to deal with the identification term bias in toxic comment classification.

\section{Model Architecture}
\label{model}
In this comparative analysis, we use two models: logistic regression and BERT. Logistic regression \cite{bishop2006pattern}, a traditional machine learning algorithm, offers simplicity and interpretability, making it a valuable baseline for comparison. On the other hand, BERT \cite{devlin}, a state-of-the-art transformer-based language model, excels in capturing complex linguistic patterns and contextual information, allowing for nuanced analysis of textual data. By employing these diverse models, we aim to comprehensively evaluate their performance, shedding light on their respective strengths and limitations in the context of our research.
\subsection{Logistic Regression Model}
Logistic regression \cite{hastie2009elements} is a widely used statistical method for binary classification tasks. Unlike linear regression, which predicts continuous outcomes, logistic regression predicts the probability of an instance belonging to a specific class. It does so by applying a logistic function to a linear combination of input features \cite{hosmer2013applied}. We implement a logistic regression model with the tf-idf vectorizer. Predictive analytic and categorization frequently make use of this kind of statistical model, commonly referred to as a logistic model \cite{bishop2006pattern}. Based on a given dataset of independent variables, logistic regression calculates the likelihood that an event will occur, such as voting or not voting. Given that the result is a probability, the dependent variable's range is 0 to 1. The posterior probability of class $C_1$ can be written as a logistic sigmoid function $\sigma(.)$ acting on a linear function of the feature vector $\phi$,

$$
p\left(C_1 \mid \boldsymbol{\phi}\right)=\sigma\left(\boldsymbol{w}^T \boldsymbol{\phi}\right) \ldots \ldots
$$
where $p\left(C_2 \mid \boldsymbol{\phi}\right)=1-p\left(C_1 \mid \boldsymbol{\phi}\right)$

For a dataset {$\phi$, tn}, where the target variable [0,1] and the basis function vector $\phi(n)$ with n = 1, …., N, the likelihood function is,
$$
\begin{aligned}
p(\boldsymbol{t} \mid \boldsymbol{w})= & \prod_{n=1}^N y_n^{t_n}\left\{1-y_n\right\}^{1-t_n} \ldots \ldots \\
& \text { where } \boldsymbol{t}=\left(t_1, \ldots, t_N\right)^T \\
& \text { and } y_n=p\left(C_1 \mid \boldsymbol{\phi}_n\right)
\end{aligned}
$$

\subsection{BERT Model Architecture}
The author, Devlin et al.\cite{devlin} proposed BERT which stands for Bidirectional Encoder Representations from Transformers based on fine-tuning approach and alleviates open GPT’s \cite{radford2019language} unidirectional constraint by using a “masked language model” (MLM) pre-training objective. It uses a famous attention mechanism called transformers which was proposed by Vaswani et al. \cite{vaswani2017attention}. The transformer consists of two blocks, an encoder and a decoder, and was designed to address issues with the current machine translation models. However, BERT only employs the transformer's 12 encoder blocks layered one on top of the other, with a 12-head multi-head attention block being used in each block. Each word embedding in an encoder block will have 12 separate attention patterns with regard to each other word in the supplied phrase. BERT-base comprises over 110 million trainable parameters in total.

\begin{figure}[htbp]
    \centering
    \includegraphics[width=0.8\textwidth]{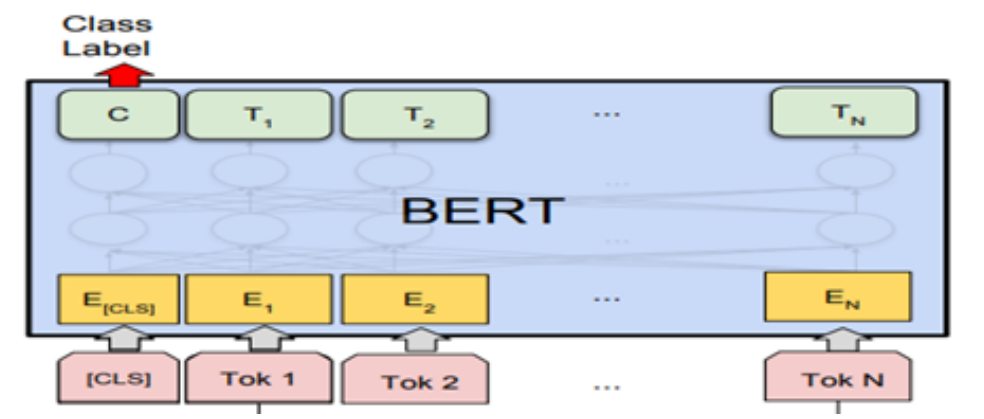} 
    \caption{Fine tuning BERT}
    \label{finetune}
\end{figure}

 Here BERT models are fine-tuned shown in Figure \ref{finetune} using the labelled data for classification tasks. The fine-tuned model is different for each task but share share the same pre-trained parameters. BERT decomposes the input sentence into WordPiece tokens that helps improve the representation of the input vocabulary and reduce its size, by segmenting complex words into subwords. These subwords can even form new words not seen in the training samples, thus making the model more robust to out-of-vocabulary (OOV) words. BERT further inserts a classification token ([CLS]) before the input tokens, and the output corresponding to this token is used for classification tasks that target the entire input. 

\section{Literature Review}
\label{liter}

\textbf{RQ1.} What metrics are best suitable to measure the bias minimization performance
of the logistics regression and Fine-tuned Bert models and why?\\

A literature review is conducted to answer the RQ1. Recent machine learning research has provided several definitions of fairness, and numerous metrics have also been proposed to evaluate the fairness of machine learning models. However, we concentrate to select metrics to  measure the fairness of the text
classification models that related to current research study.To determine the metric that best fits the problem, we conducted a literature analysis on the many metrics that were employed to assess biases in text classification models. Several metrics were offered to measure this sort of biases in the text classification models.

\subsection{Error Rate Equality Difference}
This metric is proposed by park \cite{park2018reducing} that measure of the false positive and negative equality gaps that measure the different between the true positive rates between the subgroup and the overall background as following.

$$
\begin{aligned}
F P E D=\sum_{t \in T}\left|F P R-F P R_t\right| \\ 
F N E D=\sum_{t \in T}\left|F N R-F N R_t\right|
\end{aligned}
$$

A model with a low error rate equality difference (False positive equality difference, FPED and False negative equality difference, FNED) is seen to be the fairest model when evaluating the bias mitigation of various models.
However, this measure is only applicable to classifier models with binary classes. The output of many models, however, is a probability distribution. The Equality Gap fails for many applications because it only assesses the model at one particular threshold, whereas these probability scores may be employed in other ways or with various thresholds.

\subsection{Counterfactual token fairness (CTF)}
A tractable metric proposed by \cite{garg2019counterfactual} for measuring counterfactual fairness in text classification by the following expressed function.They evaluated the CTF gaps for inputs of at most ten tokens in length for nontoxic and toxic comments separately due to asymmetric counterfactual.However, The equality gap is measured with respect to a given counterfactual generation function for a particular threshold chosen to maximize the accuracy over a designated test set and have to use a threshold to convert the
output probabilities into distinct classification outputs.

$$
\operatorname{CF~GAP}_{\Phi}(x)=\underset{x^{\prime} \sim \operatorname{Unif}[\Phi(x)]}{\mathbb{E}}\left|f(x)-f\left(x^{\prime}\right)\right|
$$

\subsection{Pinned AUC}

Author\cite{dixon2018measuring} introduces a threshold agnostic metric for unintended bias known as pinned AUC. It calculates AUC two for every identity sub group containing two equally balanced components. Combining or pinning the identity subgroup with the underlying distribution enables the AUC metric to capture
the performance difference of the model on the identity subgroup with respect to the
original distribution, there by giving a direct measure of bias. Bias metrics are expressed as 

$$
\begin{gathered}
p D_t=s\left(D_t\right)+s(D),\left|s\left(D_t\right)\right|=|s(D)| \\
p A U C_t=A U C\left(p D_t\right)
\end{gathered}
$$
However, the metric is not robust to variations in the class distribution between different identity groups. In addition, with a single metric, some important information may
be hidden as different types of bias could obscure one another mentioned by paper\cite{borkan2019limitations}

\subsection{Nuanced Metrics}
\begin{figure}[htbp]
    \centering
    \includegraphics[width=\textwidth]{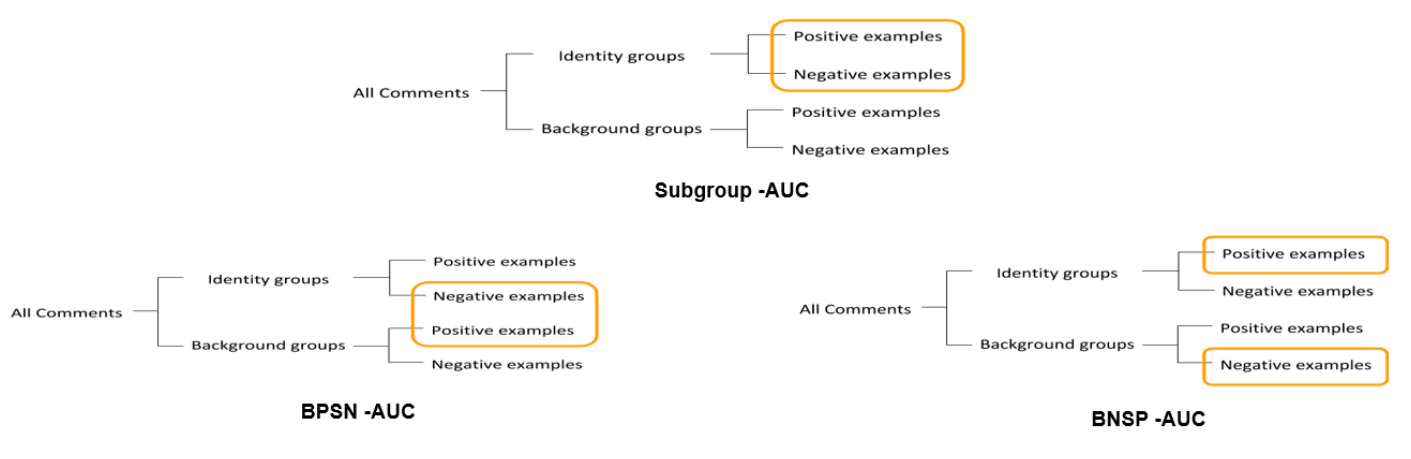}
    \caption{Nuanced Metrices}
    \label{nuance}
\end{figure}

In order to overcome the problem of pinned AUC, same author Lucas et al. \cite{borkan2019nuanced} had proposed metrics namely subgroup-AUC, BPSN-AUC, BNSP-AUC can be used in finding different kinds of biases. Dataset will be divided into two subgroups where one is related to identity groups consists of toxic and toxic comment and another Background group which contain also toxic and non toxic comment explained in Figure \ref{nuance}.

\textbf{Subgroup-AUC:}  Calculate AUC for taking identity subgroup and represents the model understanding about identity subgroup. Lower value of AUC means the model cannot identify between toxic and non toxic comment contains identity terms in the subgroup.

\textbf{BPSN-AUC:} Taking toxic comment from background group and non toxic comment from subgroup, it calculate AUC.A low AUC indicates that the model mixes identity mentions in non-toxic comments with identity mentions in toxic remarks.

\textbf{BNSP-AUC:} Measure AUC taking non toxic comment from background group and toxic comment from identity group.
Lower value of AUC will mean that the model confuses between toxic comment that mentions identity and non-toxic comments that doesn’t mentions identity.

$$
\text { Score }=\mathrm{w}_0 \mathrm{AUC} \text { overall }+\sum_{\mathrm{a}=1}^{\mathrm{A}} \mathrm{w}_{\mathrm{a}} \mathrm{M}_{\mathrm{p}}\left(\mathrm{m}_{\mathrm{s}, \mathrm{a}}\right)
$$

where:

A = number of submetrics (3)

 m= bias metric for identity subgroup  using sub-metric (subgroup-AUC, BNSP, BPSN)

 w= weighting for the relative importance of each submetric; all four  values set to 0.25 as suggested by Vaidya et al.\cite{vaidya2019empirical}

\subsection{Metrics Selection}

After studying various metrics summary in \textbf{section 4} used to measure the biases in text classification, we decided to use the subgroup-AUC, BPSN-AUC and
BNSP-AUC to measure the bias minimization performance of the model over individual
identity subgroup and the metric generalised mean AUC mentioned in above equation is used to measure the overall bias minimization performance.

\section{Experiments}
\label{exp}
\textbf{RQ2.} How well the logistic regression and Fine-tuned Bert models can measure the toxicity by reducing the unintended bias  over certain identity terms?

To answere \textbf{RQ2}, we conduct an experiment where we compare traditional machine learning with BERT model. First, we train all  models and then  test these models to compare the bias minimization performance using the metrics obtained from \textbf{RQ1} in section 4. 

\subsection{Experiment set-up / Tools}

Python was chosen for its robust support in machine learning and deep learning projects. Libraries such as PyTorch, NumPy, Sci-kit learn, SciPy, and Pandas were leveraged for technical and scientific computing tasks. Additionally, the research made use of the Google Colab hardware environment to enhance computational capabilities.

\subsection{Dataset}

The data set \footnote{https://www.kaggle.com/c/jigsaw-unintended-bias-in-toxicity-classification } we collected from kaggle and is provided by the Conversation AI team, a research initiative founded by Jigsaw. Data set contains 1.8 million observations and 45 features. The \textbf{“comment text”} column contains the comment from diverse range of conversations and “target” column indicates how toxic a comment is and target value $\ge$ 0.5 means comment is toxic. Additional toxicity sub-type attributes are severe toxicity, obscene, identity attack, insult, threat, and sexually explicit. Reaction variables are funny, wow, sad, likes, disagree. Identity variables
can be classified into five categories.

\textbf{Gender:} male, female, transgender, other gender.\\
\textbf{Sex:} heterosexual, homosexual gay or lesbian, bisexual, other sexual orientation.\\
\textbf{Religion:} Christian, Jewish, Muslim, Hindu, Buddhist, atheist, other religion.\\
\textbf{Race:} black, white, Asian, Latino, and other race or ethnicity.\\
\textbf{Disability:} physical disability, intellectual or learning disability, psychiatric or mental
illness, other disability.\\
Toxicity labels were obtained from the human raters. Each comment was rated by 10 to thousand raters.Raters marked each comment as very toxic, toxic, hard to say or no toxic. They also tried to guess the gender targeted in the comment etc. Here are some examples of the data-set.\\
\textbf{Comment:} “haha you guys are a bunch of losers.”\\
Target Label (Toxicity): 0.89\\
Severe toxicity: 0.02\\
Identity attack: 0.02\\
Insult: 0.87\\
All others: 0.0\\
\textbf{Comment:} “The woman is basically a slave.”\\
Target Label (Toxicity): 0.83\\
Identity attack: 0.83\\
Insult: 0.83\\
Female: 1.0\\
All others: 0.0\\
\textbf{Comment:} “I love the idea of upvoting entire articles.”\\
Target Label (Toxicity): 0.0\\
All others: 0.0\\

\subsubsection{Exploratory Data Analysis}

The primary objective of Exploratory Data Analysis  is to uncover the underlying structure of data and determine the trends, patterns, and relationships among variables.  EDA extracts meaningful conclusions and insights from the data, enabling a rational and strategic approach to the problem-solving process\cite{somya}. By using statistical summary and graphical representations, it can be utilized to discover trends and patterns, to detect outliers or unexpected events, and to find interesting relationships between the variables, among other things.

\begin{figure}[ht]
    \centering
    \includegraphics[width=0.46\textwidth]{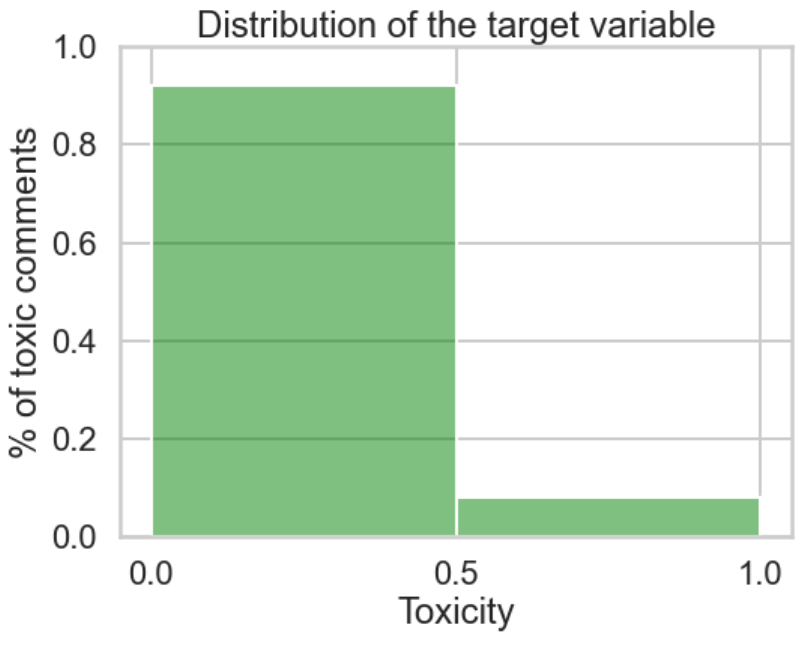}
    \caption{Distribution of the target variable}
    \label{fig1}
\end{figure}

Figure \ref{fig1}. represents the distribution of the target variable which has two categories such as non-toxic (0) and toxic (1). Target variable $<$ 0.5 belongs to non-toxic and target variable$\ge$ 0.5 belongs to toxic comments.
From Figure \ref{fig1}, we can see that the data-set is highly imbalanced. Among 1.8M observations, 92  \% (1.66M) of the data belongs to non-toxic and only 8 \% (0.14M) of the data belongs to toxic comments.

\begin{figure}[ht]
    \centering
    \includegraphics[width=\textwidth]{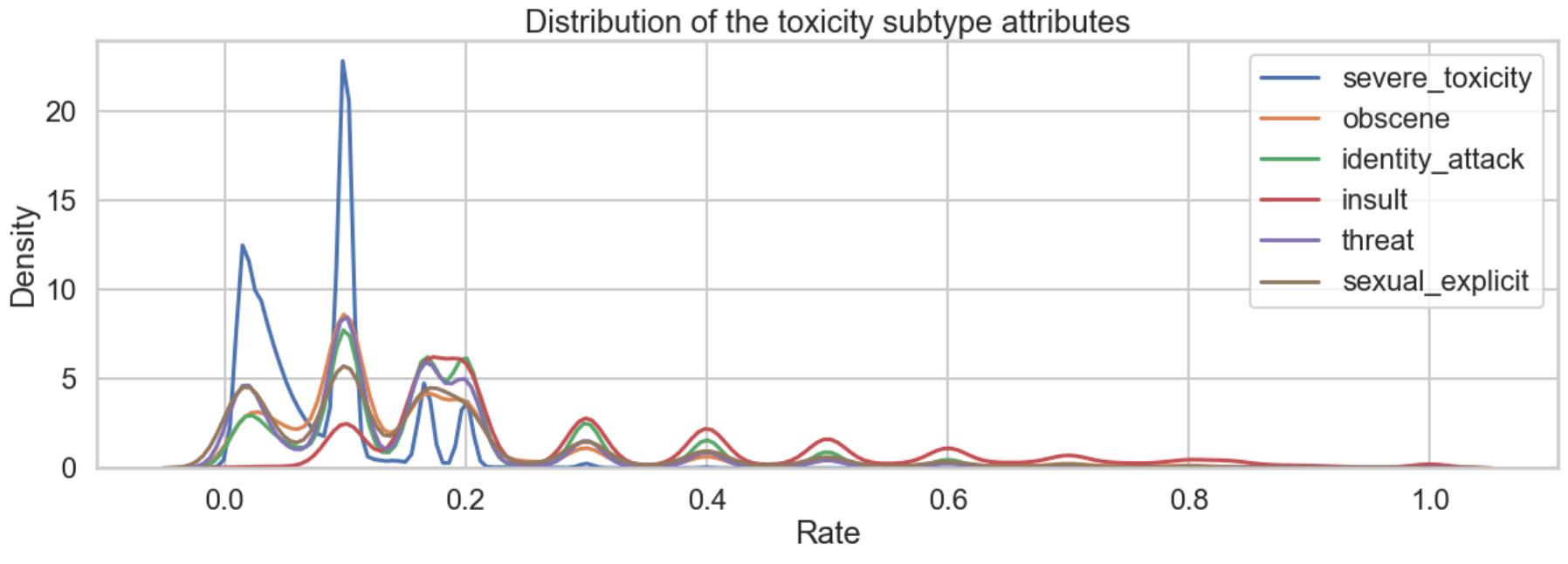}
    \caption{Distribution of the toxicity subtype attributes}
    \label{fig2}
\end{figure}

Figure \ref{fig2}, represents the distribution of the toxicity sub-type attributes such as severe toxicity, obscene (offensive or disgusting), identity attack, insult, threatened sexually explicit (sexual harassment). Most of the attributes belong to the value less than 0.5.

\begin{figure}[ht]
    \centering
    \includegraphics[width=\textwidth]{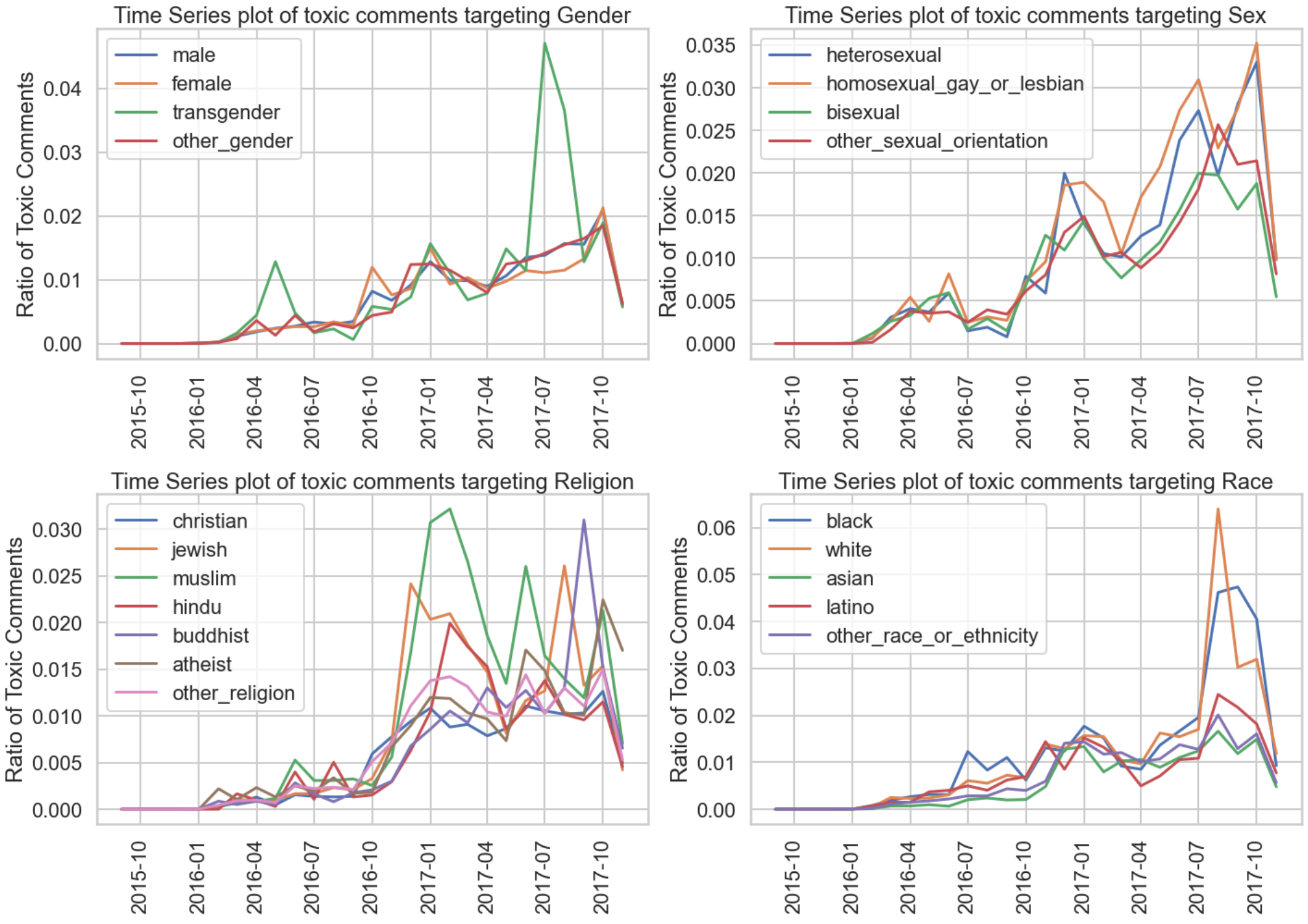}
    \caption{Time series analysis of the toxicity subtype attributes}
    \label{fig3}
\end{figure}

Figure \ref{fig3} (left upper one) represents the time series analysis of the ratio of toxic comments and gender identity variables such
male, female, transgender, and other gender from 2015
to 2017. The data-set contains news articles from
different sources. From the Figure \ref{fig3} (left upper one), we can see that
transgender people are mostly targeted in the online
toxic comments. We can also observe that there is a
sharp increase of toxic comments regarding
transgender people during July 2017 maybe because
US President Donald Trump twitted about a ban on
transgender people serving in the military. The Figure \ref{fig3} (right upper one) represents the time series analysis of the ratio
of toxic comments and sex identity variables such
homosexual gay or lesbian, heterosexual, bisexual,
and other sexual orientation. Before the shutdown of
the Civil Comments, at October 2017, the ratio of toxic
comments targeting homosexual gay or lesbian, and
heterosexual reached maximum.

Figure \ref{fig3} (left lower one) represents the time series analysis of the ratio of toxic comments and religion related identity variables such Christian, Jewish, Muslim, Hindu, Figure \ref{fig3} (lower right one) represents the time series analysis of the ratio
of toxic comments and Race identity variables such as
white, black, Asian, Latino, and other race or ethnicity.
It’s still surprising that people live in modern society
and also target human beings by their color especially
the white and black.

\begin{figure}[!h]
    \centering
    \includegraphics[width=\textwidth]{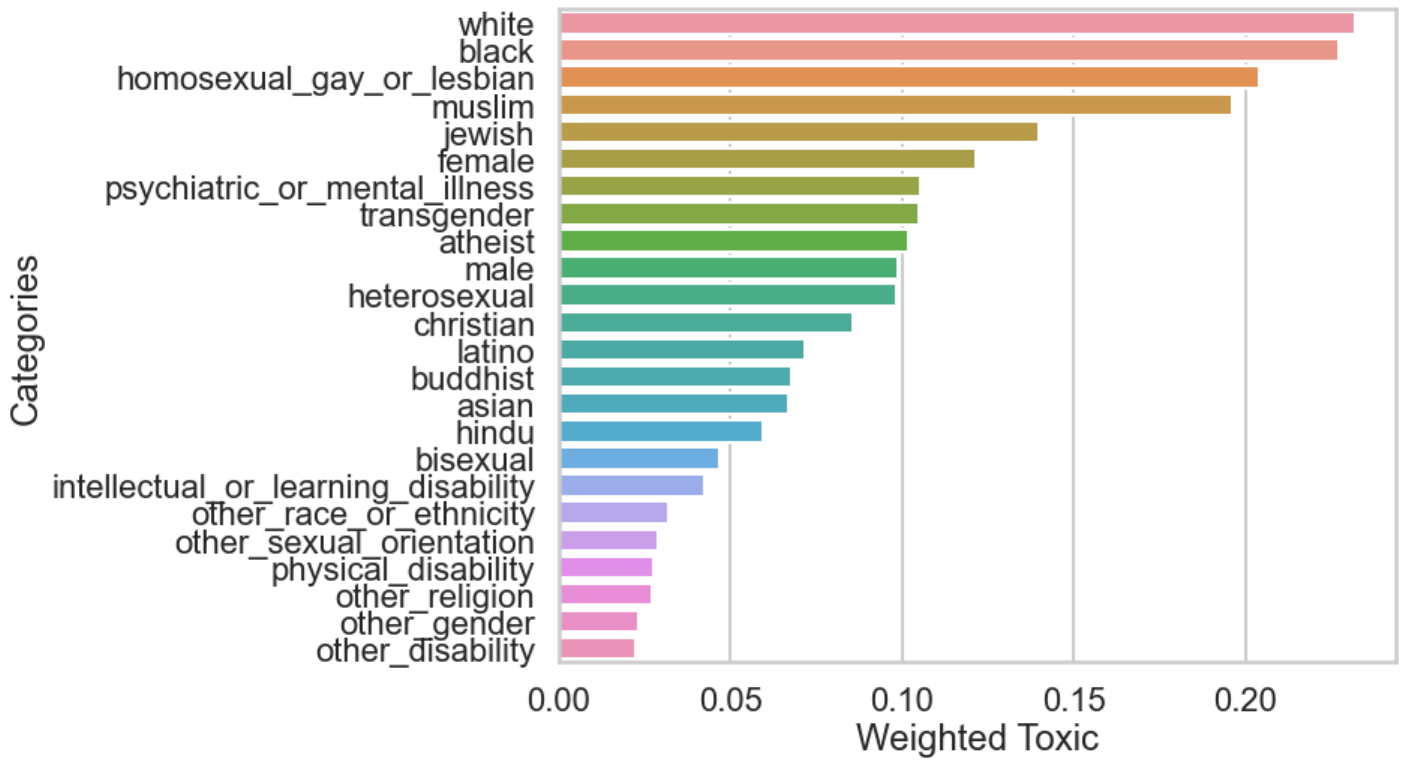}
    \caption{Percent of toxic comments related to different
identities using target and population amount of each
identity as weights}
    \label{fig4}
\end{figure}

From the Figure \ref{fig4}, among these 24 identity variables, the most targeted identities are white, black, homosexual gay or lesbian, Muslim, Jewish, female etc. Here, for each observation we have a value of target variable which represents how toxic the comment is. Each identity variable has also a value between 0 to 1 to identify how much they have been targeted. Using these two aspects, we can find which identities are more frequently related to toxic comments. To determine the weighted toxicity, first calculate the sum
of the product of each identity variable with the target variable. Then, find the ratio of this sum of product to the number of identity variables which are greater than zero. 

\begin{figure}[h]
    \centering
    \includegraphics[width=0.6\textwidth]{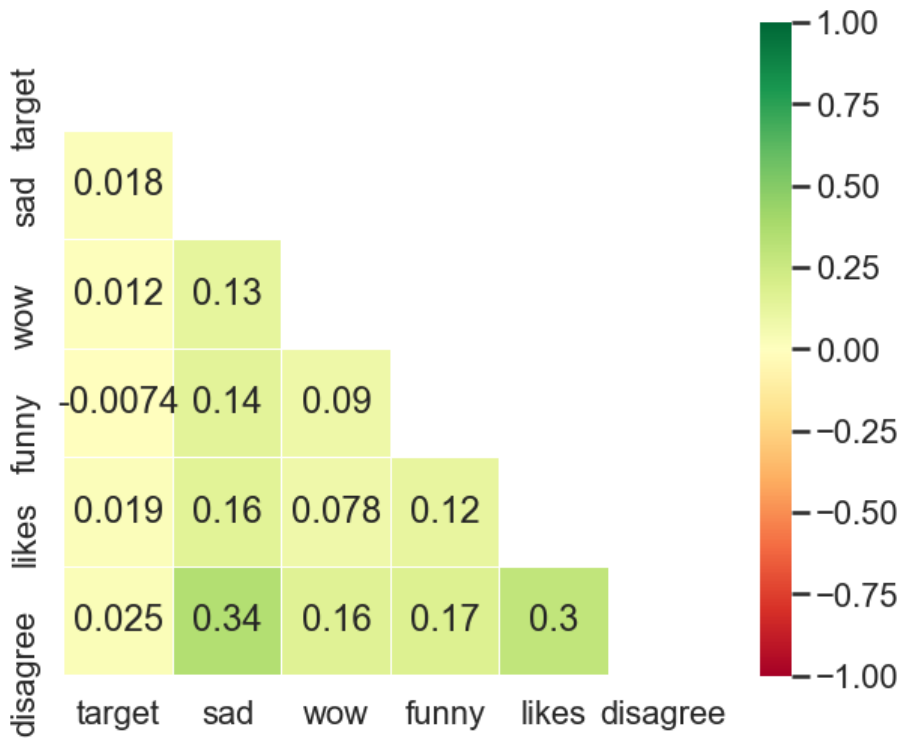}
    \caption{Correlation heatmap of target variable and the
reaction variables in the comments}
    \label{fig5}
\end{figure}

In order to understand the relation between different variables with the target variable, heat-map of correlations has been generated in two steps. First, a heat-map is created at Figure \ref{fig5} to take a look at the pairwise correlations between target variable and the reaction variables in the comments such as sad, wow, funny, likes, disagree. The strongest correlation value 0.34 exists between ‘sad’ and ‘disagree’ maybe because of the content of comments that people disagree, make them sad as well. Looking at the correlation to the target variable, the correlations are very weak. Maybe the reason is the data is collected from the Civil Comments platform from 2015 to 2017 and these data are not coming from popular social networking sites such as Reddit, Facebook, Instagram or Twitter where reactions are very popular. This can be an explanation for the low number of reaction votes and the weak correlation.

\begin{figure}[ht]
    \centering
    \includegraphics[width=0.8\textwidth]{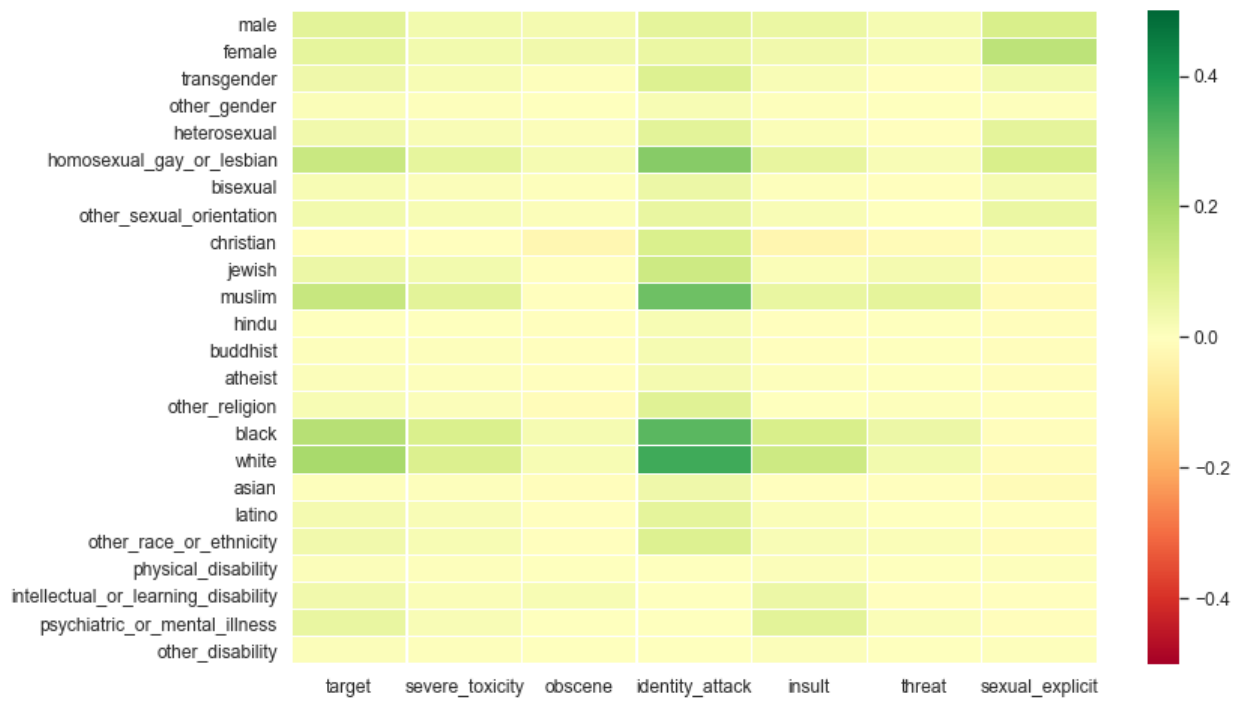}
    \caption{Correlation heat map of target variable, toxicity sub type attributes and the identity variables}
    \label{fig6}
\end{figure}

Another correlation heat-map is generated at Figure \ref{fig6} to observe the relationship between target variable,
toxicity sub type attributes and the identity variables. The sub type attribute ‘identity attack’ has higher
positive correlation value with the race white and black, religion Muslim, and sex homosexual gay or
lesbian. The exact same identity variables have high correlation with the target variable which verify the
weighted toxicity of the identity variables at Figure \ref{fig4}. Overall, the additional data, coming along with the
target variable and the comments, seems not really contributing to predict the toxicity. Thus, we decided on
excluding the information for the following prepossessing and modeling. This does not mean that
for other data sets from other platforms will lead to the
same conclusions.

\subsection{Data Preprocessing}

\begin{itemize}
    \item Before applying for predictive modeling, text data requires special preparation. We remove-html tags, punctuation, stop words and special characters like , or .,$ @$. We convert the word to lowercase, expand contractions, check if the word is made up of English letters and is not alpha-numeric and convert the word to lowercase as tasty and Tasty considered as different tokens. We also remove Stop words(e.g. a, the, am, are etc.) and expand contractions(e.g. “aren’t” expanded to “are not”).
    
    \item From Exploratory Data Analysis in section \textbf{5.2.1}, Dataset is highly imbalanced shown in Figure\ref{fig1}. Among 1.8M observations 92\% (1.66M) of the data belongs to non-toxic and only 8\% (0.14M)of the data belongs to toxic comments. We split whole dataset into 80\% as training set, 20\% as as test set and used $class_weight$ hyperparameter from Scikit Learn classifiers which contain a $class_weights$ option that can be set to "balance" or provided a custom dictionary to specify how to prioritize the significance of \textbf{unbalanced data}.This hyper parameter is fine tuned using grid search methods. We train a BERT and logistic regression model model setting class weights as balance where model will take same portion of data sample from two groups. 

    \item \textbf{TF-IDF:}  As models can’t understand text directly, We vectorize data using TF-IDF method from sci-kit learn library. Term Frequency(TF) calculates number of times that term t occurs in document d and inverse-document frequency (IDF) measures how much information the word provides by the following equation\cite{luhn1957statistical}.
  $$
\operatorname{tf}(t, d)=\frac{f_{t, d}}{\sum_{t^{\prime} \in d} f_{t^{\prime}, d}}
$$,
$$
i d f(w)=\log \left(\frac{N}{d f_t}\right)
$$
\item  \textbf{WordPiece tokenizer:} Bert use Word Piece tokenizer \cite{wu2016google} to vectorize text data that can be used from pytorch transformer package. It works by splitting words either into the full forms either takes one word becomes one token or take word into pieces where one word can be broken into multiple tokens. 
\end{itemize}

\subsection{Model Loss Function and Optimizer}
 During training process, We used cross-entropy loss \cite{zhang2018generalized} to measure the classification model’s performance whose output is a probability value between 0 and 1. The goal of this loss function is to penalize correctly classified data with a low penalty and incorrect predictions with a large penalty. Each sample's likelihood of belonging to one of the classes is calculated, and the cross-entropy between these probabilities is used as the cost function. Here is the equation to calculate cross entropy loss.
 $$
H_p(q)=-\frac{1}{N} \sum_{i=1}^N y_i \cdot \log \left(p\left(y_i\right)\right)+\left(1-y_i\right) \cdot \log \left(1-p\left(y_i\right)\right)
$$
We used Adam optimizer \cite{kingma2014adam} (adaptive moment estimation) which is an adaptive learning rate method and
computes individual learning rates for different parameters. Adam uses estimations of first moment (mean) and second moment (variance) of gradient to adapt the learning rate for each weight of the neural network.
\subsection{Model Hyper-Parameter Tuning}
A model's learning process is controlled by a hyper-parameter. The parameters' values are tuned to determine how successfully a model trains.GridsearchCV \cite{liashchynskyi2019grid}is a cross validation method to find out best parameter for traditional machine learning model. For logistic regression model, we used gridsearchCV from sklearn library to hyper-parameter tuning. We tuned the learning rate with 0.0001, with 100 value for Batch size and use one epoch for BERT model. 

\subsection{Model Evaluation Metrics}
\begin{itemize}
    \item \textbf{Classification performance:} To measure the overall classification performance of the models We have used  Overall-AUC described by the paper \cite{fawcett2006introduction} which calculate on the entire test set. AUC is the area under the ROC(Receiver Operating Characteristics) \cite{borkan2019nuanced} that obtained by plotting true positive rates(on the y-axis) and false-positive rates(on the x-axis) of the classifier. If a classifier has an AUC score close to 1 then it is doing a good job of correctly classifying positive and negative classes and if a classifier has an AUC close to 0 then it is doing a bad job of classifying
positive and negative classes.
    \item \textbf{Bias minimization performance:} For measuring the bias minimization performance over different identity subgroups We have used the Subgroup-AUC, BPSN-AUC, BNSP-AUC and Generalized mean AUC that were selected in the literature review in section \textbf{4}. 
    
\end{itemize}

\section{Experimental Result}
\label{result}
In this section, We will discuss the results obtained from the experiment. First, We will discuss the classification performance of the models i.e Overall-AUC. Finally, We will discuss the bias minimization performance of logistic regression and BERT models used in this experiment.
\begin{itemize}
    \item \textbf{Classification Performance:}The table \ref{restable} presents the model classification report based on overall-AUC.
    
\begin{table}[!h]
\centering
\caption{Overall-AUC for models}
\label{restable}
\begin{tabular}{|c|c|}
\hline

Model               & Overall-AUC \\ \hline
Logistic regression & 0.56        \\ \hline
Fine-tuned BERT     & 0.86        \\ \hline
\end{tabular}
\end{table}

  \item\textbf{Bias minimization performance:} The table \ref{logist}\ref{bert} presents the values of the metrics subgroup-AUC, BPSN-AUC, BSPN-AUC corresponding to individual identity subgroup for logistic regression and BERT model.Generalized mean AUC for these two model are in the table\ref{general}.
  
\begin{table}[!h]
\centering
\caption{Logistic regression }
\label{logist}
\begin{tabular}{|c|c|c|c|c|}
\hline
Model                                                         & subgroup                        & subgroup\_auc & bpsn\_auc & bnsp\_auc \\ \hline
\begin{tabular}[c]{@{}c@{}}Logistic\\ regression\end{tabular} & male                            & 0.56          & 0.68      & 0.42      \\ \hline
                                                              & female                          & 0.57          & 0.71      & 0.39      \\ \hline
                                                              & christian                       & 0.60          & 0.75      & 0.37      \\ \hline
                                                              & muslim                          & 0.60          & 0.72      & 0.41      \\ \hline
                                                              & white                           & 0.61          & 0.71      & 0.43      \\ \hline
                                                              & jewish                          & 0.62          & 0.77      & 0.40      \\ \hline
                                                              & black                           & 0.63          & 0.75      & 0.41      \\ \hline
                                                              & homosexual\_gay\_or\_lesbian    & 0.65          & 0.75      & 0.42      \\ \hline
                                                              & psychiatric\_or\_mental\_illnes & 0.68          & 0.71      & 0.50      \\ \hline
\end{tabular}
\end{table}

\begin{table}[!h]
\centering
\caption{BERT result}
\label{bert}
\begin{tabular}{|c|c|c|c|c|}
\hline
Model & subgroup                        & subgroup\_auc & bpsn\_auc & bnsp\_auc \\ \hline
BERT  & male                            & 0.89          & 0.88      & 0.95      \\ \hline
      & female                          & 0.898         & 0.887     & 0.956     \\ \hline
      & christian                       & 0.90          & 0.91      & 0.94      \\ \hline
      & muslim                          & 0.81          & 0.80      & 0.96      \\ \hline
      & white                           & 0.82          & 0.77      & 0.97      \\ \hline
      & jewish                          & 0.88          & 0.86      & 0.96      \\ \hline
      & black                           & 0.82          & 0.75      & 0.97      \\ \hline
      & homosexual\_gay\_or\_lesbian    & 0.79          & 0.75      & 0.96      \\ \hline
      & psychiatric\_or\_mental\_illnes & 0.90          & 0.83      & 0.97      \\ \hline
\end{tabular}
\end{table}

\begin{table}[!h]
\centering
\caption{Generalized mean AUC}
\label{general}
\begin{tabular}{|c|c|}
\hline

Model               & Generalized mean-AUC \\ \hline
Logistic regression & 0.57        \\ \hline
Fine-tuned BERT     & 0.89        \\ \hline
\end{tabular}
\end{table}

\end{itemize}

\section{Discussion}

Answer to the \textbf{RQ1:} “What metrics are best suitable to measure the bias minimization performance of the proposed model and why?” is obtained by conducting a literature review in section 4. The metrics subgroup-AUC, BPSN-AUC, BSPN-AUC, and Generalised mean AUC are chosen to evaluate the bias minimization performance in the text classification models after studying several metrics previously used to measure biases in text classification models. 

We have conducted an experiment to answer the \textbf{RQ2:} How well two model (logistic regression and BERT)can measure the toxicity by reducing the unintended bias ? 
On observing the results in the section \textbf{6}, it is clear that the classification performance of the BERT-based model outperforms that of the logistic regression model. Similar to the above, a deeper look at the per-identity metrics (subgroup-AUC, BPSN-AUC, and BNSP-AUC) and the generalized mean AUC for the two models reveals that the BERT-based model performs better at bias minimization than the Logistic Regression model. The BERT based model performed better than the other model in terms of BPSN-AUC and BNSP-AUC scores in all identity subgroups. This suggests that the BERT based model is better able to decrease false positive and false negative biases. When compared to the logistic regression model, the BERT-based model's overall bias minimization performance is better, as seen by the higher value of Generalized mean AUC.

\section{Conclusion}

Many deep learning and machine learning models have been presented out over the years to identify toxic text comments. In this  study we implement transfer learning concept on fine-tuning BERT model to built final model for reducing identity bias in toxicity classification. We also train a traditional machine learning algorithm (logistic regression) to compare with fine-tuned BERT model. After training both models, we evaluated and compared their classification performance and bias minimization performance. From the results, we can conclude that BERT based model offer better classification and Bias minimization performance when compared to the traditional machine learning algorithm like logistic regression. 

\section{Limitation and Future Research}

In this research, we used the transfer learning concept to fine-tune the BERT model, which required 23 hours to update the weights for one epoch. In the future, we want to research on  BERT model to make it compressed and computationally efficient. Another drawback is that the data set utilized in this study only presents identification labels for a portion of the comments, which is insufficient for identity information. Therefore, employing a bigger data set that incorporates comments relating to additional identity subgroups or using synthetic data preparation methods to produce more data for training the models are potential future extensions of this research work.

\section{Acknowledgement}
We would like to thank Professor Dr. Sanjay Purushotham for his great support and advising throughout this research.

\bibliographystyle{unsrtnat}
\bibliography{biblo}

\end{document}